\documentclass[10pt,twocolumn,letterpaper]{article}

\usepackage[pagenumbers]{cvpr} 

\usepackage{cuted}
\usepackage{float}
\usepackage{fontawesome}
\usepackage{graphicx}
\usepackage{multirow}
\usepackage{sidecap}
\usepackage{tabularx}

\usepackage{amsmath,amsfonts,bm}

\def\eqref#1{equation~\ref{#1}}

\def\1{\bm{1}}

\def\rvh{{\mathbf{h}}}

\def\rvx{{\mathbf{x}}}

\def\rvz{{\mathbf{z}}}

\DeclareMathAlphabet{\mathsfit}{\encodingdefault}{\sfdefault}{m}{sl}
\SetMathAlphabet{\mathsfit}{bold}{\encodingdefault}{\sfdefault}{bx}{n}

\def\gD{{\mathcal{D}}}
\def\gE{{\mathcal{E}}}

\def\gQ{{\mathcal{Q}}}

\newcommand{\method}{\textsc{RefTok}}

\definecolor{cvprblue}{rgb}{0.21,0.49,0.74}
\usepackage[pagebackref,breaklinks,colorlinks,allcolors=cvprblue]{hyperref}

\def\paperID{0000} 
\def\confName{CVPR}
\def\confYear{2025}

\title{\method: Reference-Based Tokenization for Video Generation \vspace{-1em}}

\author{Xiang Fan$^1$\thanks{Work done during internship at Amazon.}, Xiaohang Sun$^2$, Kushan Thakkar$^2$, Zhu Liu$^2$, Vimal Bhat$^2$, Ranjay Krishna$^1$, Xiang Hao$^2$ \\
$^1$University of Washington, $^2$Amazon \\
\faEnvelope\ \texttt{xiangfan@cs.washington.edu} \\
\vspace{-2em}}

\begin{document}
\maketitle
\begin{strip}
\begin{minipage}{\textwidth}
\centering
\vspace{-3em}
\includegraphics[width=\linewidth]{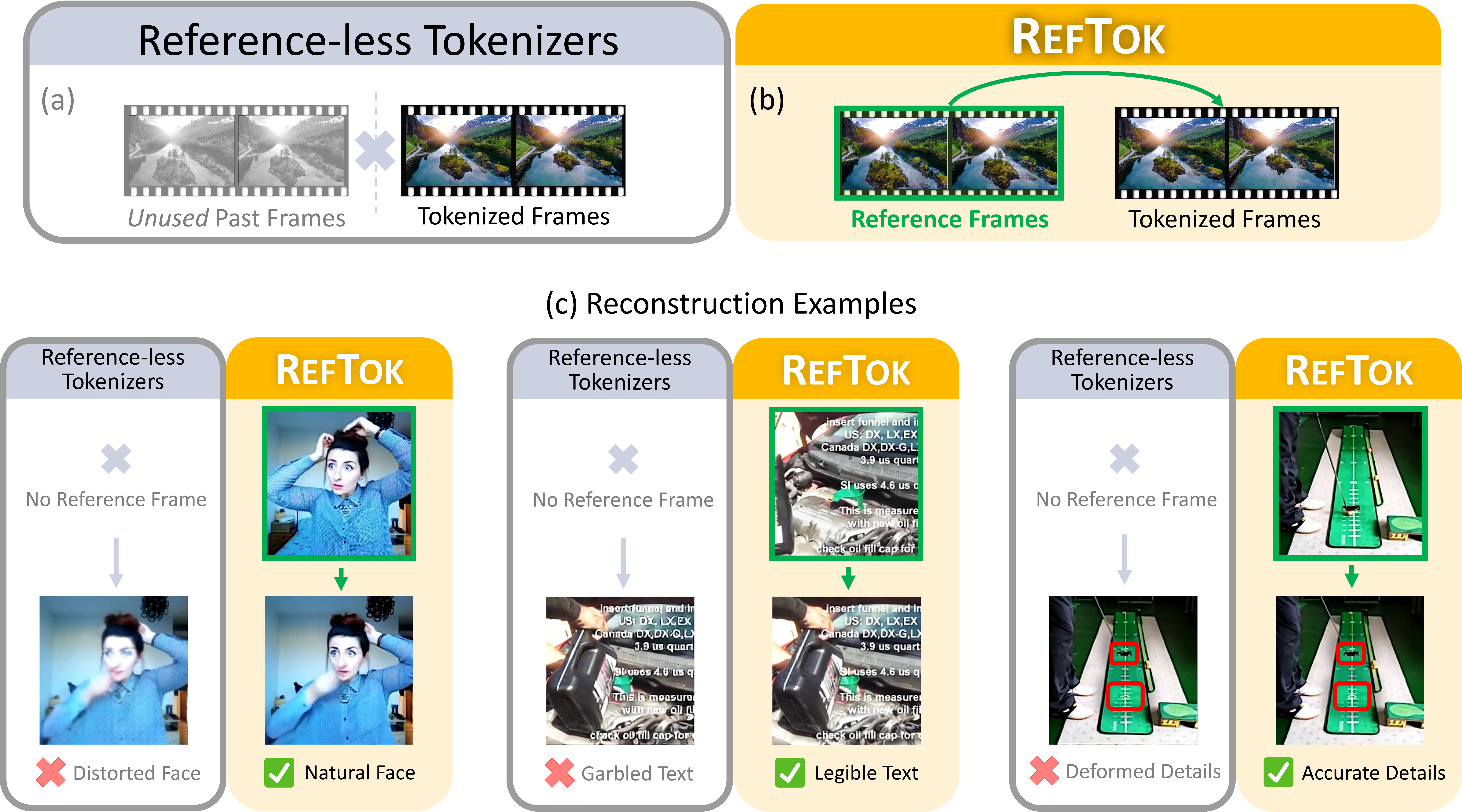}
\captionof{figure}{\method\ is a reference-based tokenizer: instead of treating each block of frames independently (shown in box (a)), \method\ allows encoding and decoding the frames conditioned on one or more reference frames (box (b)). At reconstruction time, traditional tokenizers take no reference frames while \method\ conditions on reference frames (top half of (c)). When the tokens are decoded, \method\ preserves the appearance of various objects across frames (bottom half of (c)): \method\ retains facial details, reconstructs text legibly, and preserves patterns accurately from the reference frame, in ways that reference-less tokenizers cannot preserve. As such, \method\ enables potential applications in real-world video generation tasks that require high-quality, consistent, and detailed video reconstruction.}
\label{fig:teaser}
\vspace{-0.8em}
\end{minipage}
\end{strip}

\begin{abstract}
Effectively handling temporal redundancy remains a key challenge in learning video models. 
Prevailing approaches often treat each set of frames independently, failing to effectively capture the temporal dependencies and redundancies inherent in videos. 
To address this limitation, we introduce \method, a novel reference-based tokenization method capable of capturing complex temporal dynamics and contextual information. 
Our method encodes and decodes sets of frames conditioned on an \textit{unquantized reference frame}. When decoded, \method\ preserves the continuity of motion and the appearance of objects across frames. 
For example, \method\ retains facial details despite head motion, reconstructs text correctly, preserves small patterns, and maintains the legibility of handwriting from the context.
Across $4$ video datasets (K600, UCF-101, BAIR Robot Pushing, and DAVIS), \method\ significantly outperforms current state-of-the-art tokenizers (Cosmos and MAGVIT) and improves all evaluated metrics (PSNR, SSIM, LPIPS) by an average of 36.7\% at the same or higher compression ratios.
When a video generation model is trained using \method's latents on the BAIR Robot Pushing task, the generations not only outperform MAGVIT-B but the larger MAGVIT-L, which has $4\times$ more parameters, across all generation metrics by an average of 27.9\%.
\vspace{-2em}
\end{abstract}
    
\section{Introduction}
\label{sec:intro}

Traditional video codecs such as H.264/MPEG-4 AVC~\cite{Jtc2010AdvancedVC} and AV1~\cite{rivaz2018av1} predominantly benefit from the use of \textit{reference frames} to efficiently define future frames. 
For example, H.264 allows P-frames to use patches as large as 16x16 from a previous frame as reference, and encodes a residual block to account for the difference between the reference frame and the current frame.
However, prevailing work in neural video tokenizers has largely focused on tokenizing individual blocks of frames~\cite{yu2023magvitmaskedgenerativevideo,reda2024cosmostokenizer,yan2024elastictokadaptivetokenizationimage}, which can lead to distortion and deformation of details (\Cref{fig:teaser}). This causes visual generators~\cite{rombach2022highresolutionimagesynthesislatent} to struggle to maintain temporal consistency of detailed content, such as text or human faces, even when such content is present in the context given to the generator~\cite{yu2023magvitmaskedgenerativevideo}.

To learn efficient latent representations, prevailing video tokenization models have adopted bottleneck networks such as variational autoencoders~\cite{kingma2022autoencodingvariationalbayes} and vector quantization~\cite{oord2018neuraldiscreterepresentationlearning}. 
Analogous to traditional video codecs, these methods compress video frames into a structured latent space with lower dimensionality. For example, VQ-VAE~\cite{oord2018neuraldiscreterepresentationlearning} compresses video frames into a discrete latent space with a fixed codebook size. A common setup with 8K codebook size and 16$\times$16 patch size~\cite{esser2021tamingtransformershighresolutionimage} allows only 8192 possible patches to be generated, compared to the $10^{1850}$ possibilities in RGB space. The high compression ratio sometimes presents a difficult choice: either accept that certain contents are unrepresentable (e.g. the human face and detailed text in \Cref{fig:teaser}), or increase the number of patches at the cost of increased computational complexity.

In this paper, we propose \method, a novel reference-based tokenization approach for video generation. Unlike prior methods, we don't quantize all frames. We allow reference frames to remain unquantized. All other frames are encoded and decoded relative to their reference frame. 
By allowing reference frames to skip quantization, \method \ is able to reconstruct a larger range of possible RGB space.
Our experiments will show that this formulation allows for more continuous, coherent, and detailed video reconstruction, while maintaining the same level of compression.

\method\ can be realized with some modifications to existing ViT-based tokenization architectures. Two technical insights enable \method: (1) thanks to the work on masked autoencoders, the representation encoded by an MAE encoder contains enough information to almost perfectly reconstruct the original frame; hence, we simply take a pretrained ViTMAE~\cite{he2021maskedautoencodersscalablevision} encoder and pass the reference frame directly to the decoder, skipping the quantization layer; (2) the remaining tokens can be encoded by the same ViTMAE encoder, quantized conditioned on the reference frame, and finally, the decoder can be fine-tuned to reconstruct from quantized tokens. By correctly setting the attention mask in the encoder, we can reuse the same encoder and decoder for both reference and target frames, keeping the compute budget low while maintaining proper information flow.

A key difference between \method\ and causal video encoders~\cite{yu2024languagemodelbeatsdiffusion,villegas2022phenakivariablelengthvideo} is that \method\ gives the decoder full access to the reference frame without information loss. This allows the decoder to reconstruct the target frames with the help of the ground truth reference frames, which is particularly useful for improving generation quality.

Our experiments show that \method\ produces higher quality reconstructions than state-of-the-art video tokenizers across four video datasets (K600, UCF-101, DAVIS, BAIR Robot Pushing), outperforming Cosmos~\cite{reda2024cosmostokenizer} and MAGVIT tokenizer~\cite{yu2023magvitmaskedgenerativevideo} by an average of 36.7\% in all metrics. \method\ retains facial details, preserves fine patterns, and reconstructs text accurately from the reference frame. Additionally, \method\ enables video generation models to outperform MAGVIT-L with only a MAGVIG-B-sized model on BAIR Robot Pushing video prediction by 27.9\% on average, producing higher quality videos. \method\ is also 2.29x faster and 2.38x more memory efficient than baselines at the same compression ratio.

The purpose of \method\ is not to replace reference-less tokenizers, but to introduce a new design dimension---reference frames---that is often overlooked in the designed interface of video tokenization. \method\ aims to bridge the gap between traditional video codecs and neural video tokenizers, enabling exciting potential applications that require high-quality, consistent, and detailed video reconstruction.

\section{Related work}
\label{sec:related}

Our work is based on work from discrete video tokenization and video generation, which we review in this section.

\vspace{0.2em}
\noindent\textbf{Image encoders.} Variational autoencoders (VAEs) are introduced to learn latent-space probabilitics models. VQ-VAE~\cite{oord2018neuraldiscreterepresentationlearning} introduces a vector-quantized discrete latent space to address the issue of posterior collapse and improve the quality of image generation. VQGAN~\cite{esser2021tamingtransformershighresolutionimage} extends VQ-VAE to generate high-resolution images by using CNN encoders and adversarial loss functions. In the domain of image generation, latent diffusion models (LDMs) take advantage of the encoded latent space and have shown promising results in generating high-quality images~\cite{rombach2022highresolutionimagesynthesislatent}.

\vspace{0.2em}
\noindent\textbf{Discrete video tokenization.}
Recent video tokenization techniques build on VQ-VAE~\cite{oord2018neuraldiscreterepresentationlearning} and VQGAN~\cite{esser2021tamingtransformershighresolutionimage}. MAGVIT~\cite{yu2023magvitmaskedgenerativevideo} employs a 3D tokenizer for efficient video encoding, and MAGVIT v2~\cite{yu2024languagemodelbeatsdiffusion} introduces a causal 3D CNN for better causality. ElasticTok~\cite{yan2024elastictokadaptivetokenizationimage} offers efficient adaptive tokenization for long videos. Recent tokenizers~\cite{yan2024elastictokadaptivetokenizationimage,villegas2022phenakivariablelengthvideo,reda2024cosmostokenizer} increasingly utilize the ViT architecture~\cite{dosovitskiy2021imageworth16x16words}, enabling transformer-based models for video generation. OmniTokenizer~\cite{wang2024omnitokenizerjointimagevideotokenizer} proposes joint image and video tokenization to enhance video quality.

\vspace{0.2em}
\noindent\textbf{Video generation.} Generative Adversarial Networks (GANs) have been proposed as a backbone for generative video models~\cite{skorokhodov2022styleganvcontinuousvideogenerator,acharya2018highresolutionvideogeneration,brooks2022generatinglongvideosdynamic,clark2019adversarialvideogenerationcomplex,Gupta_2022_CVPR_RVGAN,babaeizadeh2021fitvidoverfittingpixellevelvideo,saito2017temporalgenerativeadversarialnets,tulyakov2017mocogandecomposingmotioncontent,yu2022generatingvideosdynamicsawareimplicit}. However, the traditional GAN architecture is limited to generating short videos with low resolution. Recent developments in video generation have focused on video diffusion models~\cite{blattmann2023stablevideodiffusionscaling,singer2022makeavideotexttovideogenerationtextvideo, blattmann2023alignlatentshighresolutionvideo, kondratyuk2024videopoetlargelanguagemodel,menapace2024snapvideoscaledspatiotemporal,Mullan_Hotshot-XL_2023,wang2023modelscopetexttovideotechnicalreport,chen2024videocrafter2overcomingdatalimitations,wang2023videocomposercompositionalvideosynthesis,zhang2023show1marryingpixellatent}.

While diffusion-based models generate high-quality videos, they require many diffusion steps, making them computationally expensive. Discrete video tokenization methods enable efficient video generation using BERT-like models~\cite{yu2023magvitmaskedgenerativevideo,villegas2022phenakivariablelengthvideo,reda2024cosmostokenizer}. For instance, MAGVIT~\cite{yu2023magvitmaskedgenerativevideo} incorporates MaskGIT~\cite{chang2022maskgitmaskedgenerativeimage} to address autoregressive inefficiencies. Phenaki~\cite{villegas2022phenakivariablelengthvideo} uses masked transformers for text-to-video generation of arbitrarily long videos. LARP~\cite{wang2024larptokenizingvideoslearned} introduces holistic queries to enhance generation performance.

Recent work suggests that discrete tokenization may not be essential for image generation, and lightweight diffusion modules with continuous tokenization can achieve similar quality~\cite{li2024autoregressiveimagegenerationvector}. However, both methods require a bottleneck layer to regulate the latent space. While continuous tokenization could potentially be applied to video generation, our work focuses on discrete tokenizers, which have shown promising results. Replacing the discrete quantizer with a continuous bottleneck (e.g., VAE) would yield a comparable method in the continuous space.

\begin{figure*}[h!t]
    \centering
    \includegraphics[width=\textwidth]{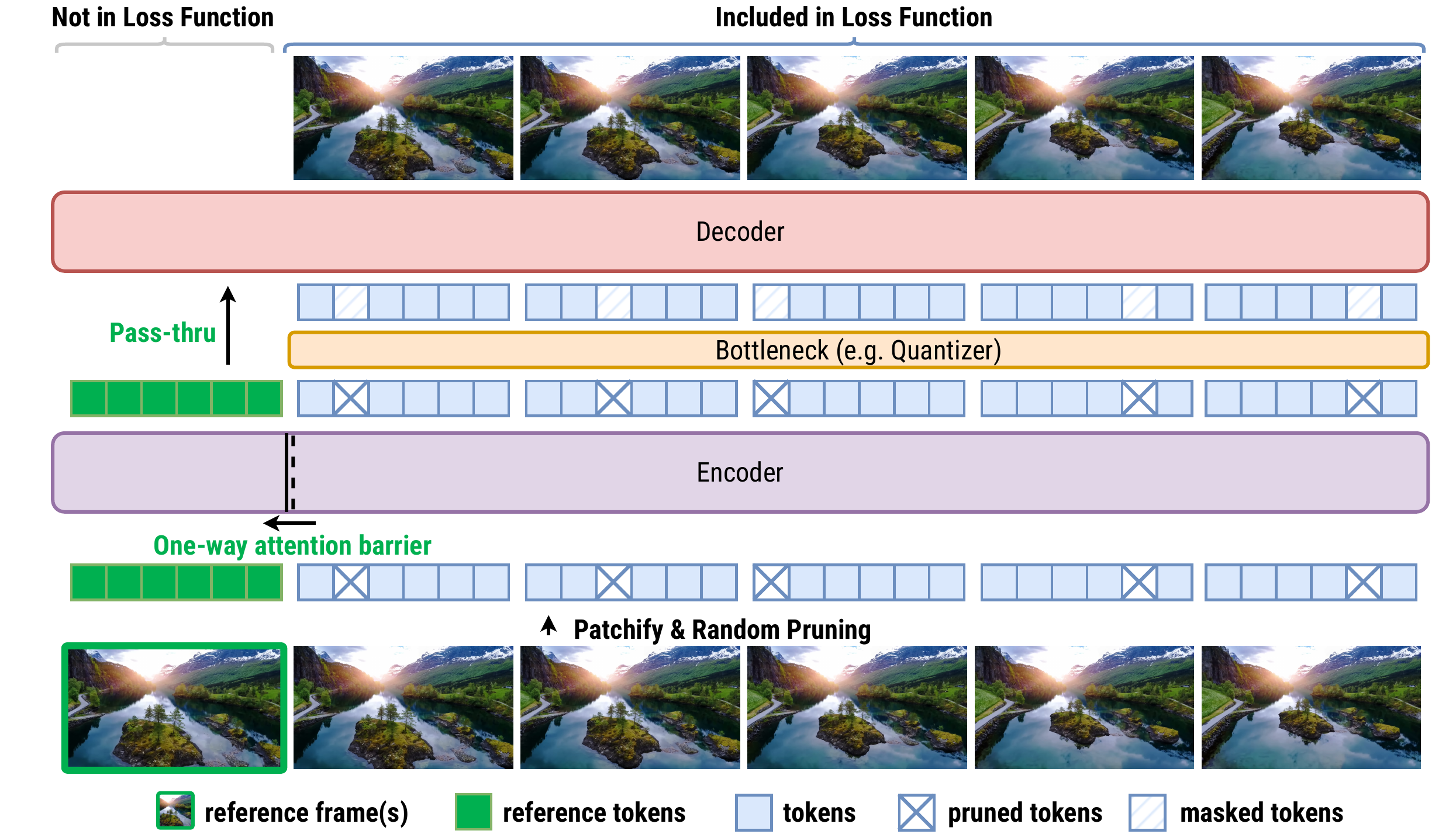}
    \vspace{-10pt}
    \caption{\method\ encodes the target frames ({\textbf{\color[HTML]{7091c0}blue}}) conditioned on reference frames (\textbf{{\color[HTML]{00b050}green}}), and allows the encoded reference frames to bypass the bottleneck layer as a conditioning signal for the decoder, allowing for higher-quality, more detailed and coherent reconstructions.}
    \label{fig:method}
    \vspace{-5pt}
\end{figure*}

\section{Method}
\label{sec:method}

We present \method, a reference-based tokenization method for video generation. At a high level, \method\ allows some tokens from a frame to be designated as reference tokens, and lets them circumvent the quantization layer; all other frames are quantized conditioned on their respective reference tokens, allowing us to compress videos at a higher compress rate while reconstructing them more accurately.

Our method leverages reference frames to supplement the information deficiency in the latent space. There are two issues with the common reference-less approach: (1) the encoder attempts to learn a distribution $p(\rvz | \rvx)$, for an input $\rvx$ and a latent $\rvz$, subject to an extremely low dimensionality for $\rvz$ that is intractable to represent the entire manifold of $\rvx$; (2) the decoder is tasked to reconstruct the original video $p(\rvx | \rvz)$ from the low dimensional latent space, which is often insufficient to capture the full complexity of the video. For example, in \Cref{fig:teaser}, the reference-less approach struggle to reconstruct text legibly.

Our method addresses these issues by allowing reference frame(s) $\rvx_r$ to precede the target frame(s) $\rvx_t$. We posit that $\rvx_r$ and $\rvx_t$ share a high amount of mutual information, as evidenced by the temporal dependencies in videos~\cite{Jtc2010AdvancedVC}. To address the information deficiency in the latent space, we let both the encoder and the decoder condition on $\rvx_r$; the encoder now learns a distribution $p(\rvz_t | \rvx_t, \rvx_r)$, for which a valid representation is only the temporal difference between $\rvx_r$ and $\rvx_t$; the decoder learns $p(\rvx_t | \rvz_t, \rvx_r)$, so that the information deficiency in the latent space is supplemented by the rich information in $\rvx_r$. This allows our tokenizer to reconstruct more detailed and coherent videos. We summarize our method in \Cref{fig:method}.

\subsection{Tokenization for video generation}

A video is typically patchified~\cite{dosovitskiy2021imageworth16x16words} before used in a generative model. For example, a 16-frame video of $256 \times 256$ dimensions has a shape of $16 \times 256 \times 256$ (ignoring the color channel dimension). A typical 3D tokenization scheme patchifies this video into a sequence of cuboids; each cuboid of size $4 \times 16 \times 16$ becomes one token, resulting in a total of $4 \times 16 \times 16$ latent tokens. In general, we consider a video of $T$ frames, each of size $H \times W$, and token patches of size $t \times h \times w$, which results in a total of $\frac T t \times \frac H h \times \frac W w$ tokens.

Let $\tau = \frac T t, \eta = \frac H h, \omega = \frac W w$. We denote our tokens as $\rvx = \{x_{i,j,k}\}$, where $1 \leq i \leq \tau, 1 \leq j \leq \eta, 1 \leq k \leq \omega$.

Typically, the tokenization model consists of an encoder-bottleneck-decoder architecture, where the encoder $\gE$ processes the all the patchified tokens, the bottleneck $\gQ$ compresses the tokens into a structured space (such as quantization~\cite{oord2018neuraldiscreterepresentationlearning}), and the decoder $\gD$ reconstructs the bottlenecked tokens into raw video frames.

Let $\rvx$ be the input tokens, $\rvh$ be the pre-bottleneck tokens, $\rvz$ be the post-bottleneck tokens, and $\hat\rvx$ be the reconstructed tokens. The process can be summarized as follows:
\begin{align}
    \rvh = \gE(\rvx) \qquad \rvz = \gQ(\rvh) \qquad \hat\rvx = \gD(\rvz)
\end{align}
$\rvz$ is used as the video representation to train the downstream video generation model.

\subsection{Tokenization for conditional video generation}

Conditional video generation is the task of generating video tokens conditioned on some information, including text or previous tokens. For applications such as future frame prediction, frame interpolating, and video inpainting~\cite{yu2023magvitmaskedgenerativevideo}, tokens are generated conditioned on a partial observation of the video.
We situate our work within this conditional tokenization paradigm.

Conditional video generation can be formulated as follows: conditioned on a given set of reference tokens $\rvx_r$, we want to generate a set of target tokens $\rvx_t$. Tokenization for conditional video generation is typically performed on the entire sequence of tokens, including both reference and target tokens:
\begin{align}
    \rvh = \gE([\rvx_r; \rvx_t]) \quad \rvz = \gQ(\rvh) \quad [\hat\rvx_r; \hat\rvx_t] = \gD(\rvz)
\end{align}
From the tokenizer's perspective, this setup is equivalent to tokenizing the entire sequence of tokens. The reconstruction loss is computed on both the reference and target tokens, i.e. $\ell([\rvx_r; \rvx_t], [\hat\rvx_r; \hat\rvx_t])$.

However, most video frames have a high degree of temporal redundancy with the reference frame. \method\ leverages this redundancy to improve the quality of the target tokens by allowing the reference tokens to bypass the bottleneck layer.

\subsection{\textbf{\fontsize{11}{\baselineskip}\selectfont\method}}

Recall that our goal is to learn an encoder for $p(\rvz_t | \rvx_t, \rvx_r)$ and a decoder for $p(\rvx_t | \rvz_t, \rvx_r)$. Taken naively, this implies that the reference frame $\rvx_r$ is fed into both the encoder and the decoder; thus both the encoder and the decoder needs to internally encode the reference frame into their latent spaces. This is inefficient, as the pre-bottleneck representation $\rvh_r$ from the encoder is already a high-quality representation of the reference frame, and the decoder should not need to encode the reference frame again. Hence, we let the decoder model learn $p(\rvx_t | \rvz_t, \rvh_r)$ instead.

We modify the tokenizer architecture in two ways: (1) reference tokens bypasses the bottleneck layer; (2) the decoder conditions on the reference tokens. The process is as follows:
\begin{align}
    \rvh_r, \rvh_t = \gE(\rvx_r, \rvx_t) \quad \rvz_t = \gQ(\rvh_t) \quad \hat\rvx_t = \gD(\rvz_t | \rvh_r)
\end{align}
This allows the reference tokens to be preserved in the continuous space through $\rvh_t$, while the target tokens are compressed and reconstructed based on the reference tokens. This architecture enables the model to leverage the higher-quality reference tokens to improve the reconstruction of the target tokens. The reconstruction loss is computed only on the target tokens, i.e. $\ell(\rvx_t, \hat\rvx_t)$, using reference tokens only as a conditioning signal.

\vspace{0.2em}
\noindent\textbf{Updates to the encoder architecture.}
To facilitate the reference-based tokenization, we introduce a few updates to the encoder architecture. Prior work has favored CNN-based architecture over transformers for encoder models~\cite{yu2024languagemodelbeatsdiffusion}. However, we find that transformers are more suitable for our task due to their ability to model long-range dependencies between the reference and target tokens. Specifically, the ViTMAE architecture contains an encoder and a decoder, which suits our needs. We initialize our autoencoder with the ViTMAE model~\cite{he2021maskedautoencodersscalablevision}, inflating 2D dimensions to 3D and expanding its sinusoidal position embeddings to accommodate the longer temporal dimension.

\vspace{0.2em}
\noindent\textbf{Preserving causality.}
To ensure causality, the encoded reference frame $\rvh_r$ must not contain information about the target frames $\rvx_t$. This prevents information from bypassing the bottleneck quantizer, preserving the utility of the quantized tokens. For tasks like frame prediction, the model should only access past frames when reconstructing future frames. We use an attention mask~\cite{vaswani2023attentionneed} within the encoder to enforce a one-way barrier between $\rvx_r$ and $\rvx_t$, ensuring no information leakage from target tokens.

\vspace{0.2em}
\noindent\textbf{Interaction with 3D tokenizers.}
3D tokenization schemes compress video tokens in both spatial and temporal dimensions. However, it is often desirable to condition on frames less than the 3D tokenization granularity. For example, a compressed cuboid of shape $4 \times 16 \times 16$ allows conditioning on 4 frames at a time, but it is often desirable to condition on only 1 frame (e.g. for image-to-video generation). For this use case, we pad $x_r$ with replicate padding to match the 3D tokenization granularity.

\vspace{0.2em}
\noindent\textbf{Addressing reference posterior collapse.}
Posterior collapse occurs when the decoder ignores the latent representation $\rvz_t$ due to weak signals from the latents. This is problematic with reference frames, as replicating the reference frame $\rvx_r$ can yield low reconstruction loss without using latents, leading to \textit{reference posterior collapse}. To mitigate this, we suggest: (1) Training with a higher frame interval than the target framerate, which improves codebook utilization by reducing frame similarity. (2) Employing codebook splitting~\cite{linde1980algorithmvectorquantizerdesign} to enhance codebook usage during early training when $\rvz_t$ is weak. Without it, many codes map to ``do nothing,'' simply replicating the previous frame. (3) While methods like FSQ~\cite{mentzer2023finitescalarquantizationvqvae,yu2024languagemodelbeatsdiffusion} snap latents to a predefined discrete space, we find codebook splitting more effective in preventing reference posterior collapse.

\vspace{0.2em}
\noindent\textbf{Token masking and pruning.} We adopt the ViTMAE architecture's approach of randomly pruning tokens during encoding and masking tokens during decoding~\cite{he2021maskedautoencodersscalablevision}. Specifically, we randomly mask and prune up to 75 tokens during training to encourage robust representation learning and maintain compatibility with adaptive tokenization schemes like ElasticTok~\cite{yan2024elastictokadaptivetokenizationimage}.

\vspace{0.2em}
\noindent\textbf{Training details.} We train on 16 A100 GPUs with a batch size of 128 using reconstruction, perceptual, and adversarial losses. See Appendix for details.

\section{Experiments}
\label{sec:experiments}

We demonstrate how \method\ enables tokenizers to take advantage of reference frames in ways existing reference-less methods do not. The key takeaways from our experiments are as follows:
(1) \method\ successfully utilizes information from reference conditioning tokens to reconstruct target pixels (36.7\% improvement).
(2) Compared to \method, existing methods demonstrate clear limitations in maintaining information from reference tokens.
(3) Thanks to its reference-based representation, \method\ enables downstream generation models to achieve state-of-the-art performance (27.9\% improvement on BAIR).
(4) \method\ is competitive in terms of speed (2.29x faster) and memory efficiency (2.38x better) compared to baseline methods, making it a practical choice for real-world applications.

\begin{figure*}[ph!]
    \centering
    \includegraphics[height=.96\textheight]{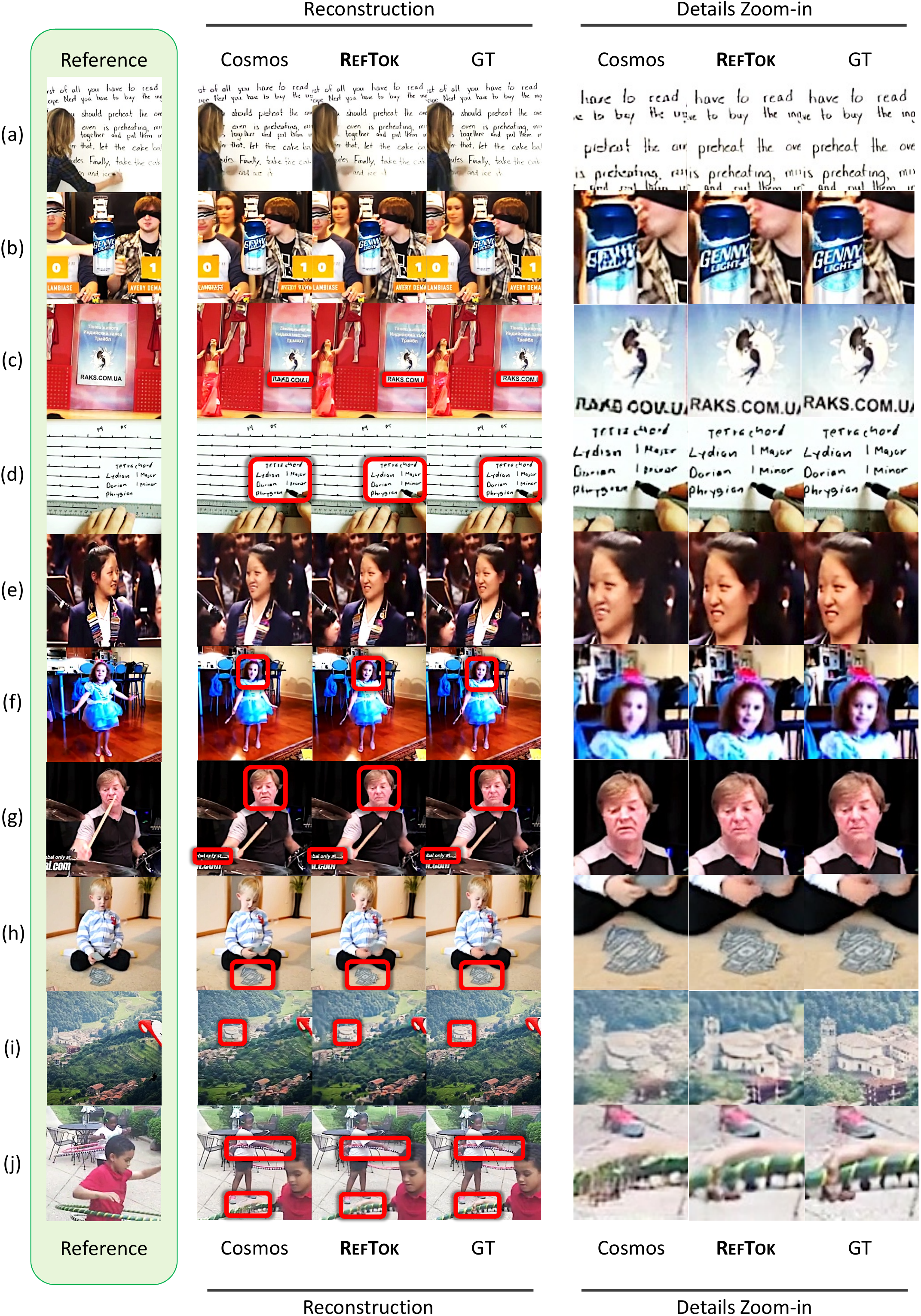}
    \caption{Reconstruction results of \method. Our method is capable of reconstructing a wide variety of concepts that are present in the reference frame, including text, human faces, detailed objects, and small patterns, even with movements and angle changes.}
    \label{fig:reconstruction}
\end{figure*}

\subsection{Datasets}
We conduct our experiments on four datasets: BAIR Robot Pushing (BAIR)~\cite{ebert2017selfsupervisedvisualplanningtemporal}, Kinetics-600 (K600)~\cite{carreira2018shortnotekinetics600}, UCF-101 (UCF)~\cite{soomro2012ucf101dataset101human}, and DAVIS~\cite{perazzi2016davis}. BAIR features robotic arm video sequences with significant temporal redundancies, ideal for evaluating \method's reference-based tokenization. K600 and UCF-101 are large-scale datasets containing 600 and 101 action categories, respectively, suitable for assessing \method's performance when referencing moving objects. DAVIS is a high-quality video dataset with 50 different objects and scenes, providing a robust benchmark for \method\ for high quality moving scenes.

\subsection{Data processing}
We preprocess the video frames by sampling 16-frame clips and resizing them to $256 \times 256$ pixels. We use a patch size of $4 \times 16 \times 16$ pixels for all experiments, unless otherwise specified. Note that the patch size is important for tokenization because it determines the granularity of tokens and therefore the compression ratio. A smaller patch size results in more tokens and higher quality videos, but also higher computational cost.

\subsection{Qualitative analysis of reconstruction}
We demonstrate the inter-frame referencing capabilities of \method\ in \Cref{fig:reconstruction}. \method\ is capable of: (1) reconstructing text almost perfectly from reference frames (\Cref{fig:reconstruction}(a)-(d), \Cref{fig:teaser}(c)-middle) despite background shift (\Cref{fig:teaser}(c)-middle) and horizontal motion (\Cref{fig:reconstruction}(c)); (2) reconstructing human faces from reference frames while avoiding uncanny valley artifacts (\Cref{fig:reconstruction}(e)-(g), \Cref{fig:teaser}(c)-left), despite head motion (\Cref{fig:teaser}(c)-left) and reference-frame occlusion (drumstick in \Cref{fig:reconstruction}(g)); (3) reconstructing fine-grained details (\Cref{fig:reconstruction}(h)-(j), \Cref{fig:teaser}(c)-right) such as the face of a dollar bill (\Cref{fig:reconstruction}(h)), doors on a far-away moving building (\Cref{fig:reconstruction}(i)), and slanted patterns on a hula hoop (\Cref{fig:reconstruction}(j)).

Comparing our method against a reference-less state-of-the-art baseline Cosmos~\cite{reda2024cosmostokenizer} in \Cref{fig:reconstruction}, we can see that Cosmos demonstrates worse text reconstruction, facial distortion, and struggles with common fine-grained patterns because it cannot use any reference-frame information. Note that Cosmos introduces 3D Haar wavelet, causal residuals, spatio-temporal convolution and attention~\cite{reda2024cosmostokenizer} to achieve its state-of-the-art performance in video tokenization, while our method is only based on the vanilla VQGAN. \textit{The introduction of reference frames alone is sufficient to outperform Cosmos in terms of reconstruction quality} under many common scenarios. We hypothesize that the performance of \method\ can be further improved by incorporating the aforementioned techniques in the future.

\subsection{Quantitative comparison of reconstruction}
We evaluate the performance of \method\ on all our evaluation datasets. We compare our method against several state-of-the-art tokenizers including MAGVIT~\cite{yu2023magvitmaskedgenerativevideo}, Cosmos~\cite{reda2024cosmostokenizer}, CogVideoX~\cite{yang2024cogvideoxtexttovideodiffusionmodels}, VideoGPT~\cite{yan2021videogptvideogenerationusing}, and OmniTokenizer~\cite{wang2024omnitokenizerjointimagevideotokenizer}. For reference-based encoding, we use the first frame as the reference frame. We maintain the same comparession ratio of 1024:1 for our model and choose the closest compression ratio that is available for the baselines. We train our model on each dataset's training split except for DAVIS; due to the small size of DAVIS, we train on Kinetics and evaluate on DAVIS.

\begin{figure}
    \centering
    \includegraphics[width=\linewidth]{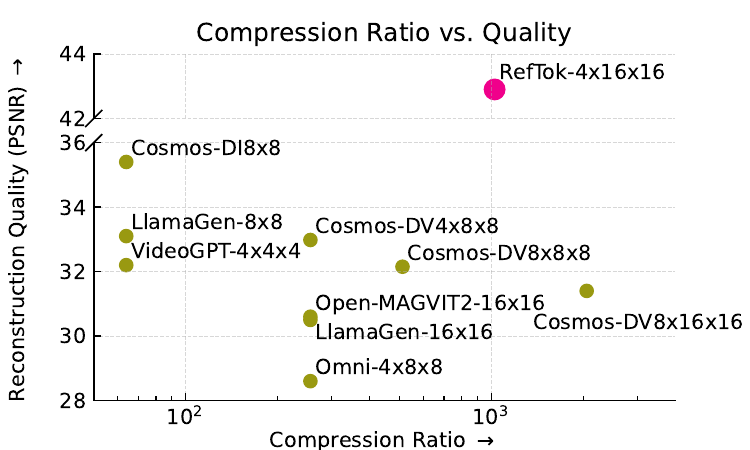}
    \caption{Comparison of compression ratio vs.\ recons.\ quality on a set of VQGAN-style tokenizers. \method\ pushes forward the Pareto front of discrete tokenizers, thanks to reference frames.}
    \label{fig:exp_compression_vs_quality}
\end{figure}

We report the quantitative results in \Cref{tab:reconstruction_results}. From the results, we can see that \method\ outperforms all baselines in cases when the compression ratio from the baseline is the same as ours, and is competitive in cases when the baseline compression ratio is lower. For example, \method\ improves PSNR and SSIM by 34\% and 14\%, respectively, and reduces LPIPS by 62\%. Overall, \method\ pushes the Pareto front of video tokenization forward by a large margin (\Cref{fig:exp_compression_vs_quality}). This demonstrates \method's ability to utilize reference frames to improve video reconstruction quality.

\vspace{.2em}
\noindent\textbf{Speed and memory efficiency.} We measure the average speed and memory efficiency of \method\ against Cosmos and MAGVIT-B in 50 repeated trials. All measurements are performed on A100 with a batch size of 8. We note that \method\ is 3.12x faster and 2.87x more memory efficient than Cosmos, and 1.46x faster and 1.89x more memory efficient than MAGVIT-B. We attribute the improvements to the use of pure transformer architecture, which downsamples the video frames to patches in a single step, instead of the multiscale downsampling of convolution.

\subsection{Training generation models with \textbf{\fontsize{11}{\baselineskip}\selectfont\method}}

\begin{figure*}[th!]
    \centering
    \includegraphics[width=.96\linewidth]{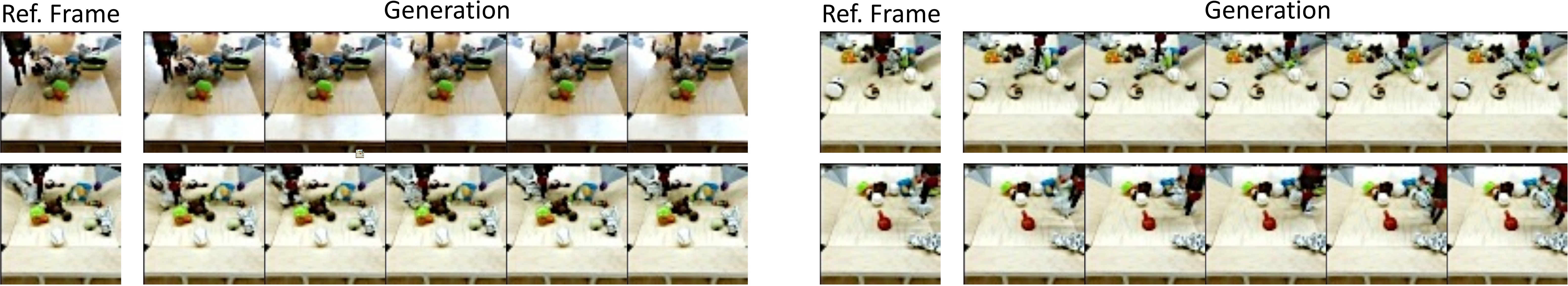}
    \caption{Results from a downstream video generation model trained on the BAIR Robot Pushing dataset with \method\ tokens. The model learns the motion dynamics without distorting the objects from the reference frame.}
    \vspace{-.8em}
    \label{fig:generation}
\end{figure*}

While the primary focus of this work is on tokenization, we also demonstrate the effectiveness using \method\ to improve downstream generation models. We use the BAIR Robot Pushing dataset, which contains a significant amount of temporal redundancies while featuring constant movement of repeating objects. We train a MAGVIT-style generation model of the BAIR dataset using the same hyperparameters~\cite{yu2023magvitmaskedgenerativevideo} and show the generated videos in \Cref{fig:generation}. As we can see, our generation model learns the motion dynamics while maintaining the shape and texture of the moving robotic arm and unmoved objects.

For quantitative evaluation, we show the results of the trained generative model in \Cref{tab:generation_results}. \method\ outperforms not only the MAGVIT-B base model, but also the large variation, MAGVIT-L, which has almost 4x more parameters, in all video evaluation metrics. For example, \method\ reduces FVD by 8\% and increases PSNR by 22\% compared to MAGVIT-L. This demonstrates that \method\ can be used to improve the performance of downstream generators.

\begin{table}[t]
    \centering
    \scriptsize
    \begin{tabularx}{\linewidth}{l | @{\hspace{.6em}} *{1}{>{\centering\hsize=5em}X} @{\hspace{1em}} *{1}{>{\centering\arraybackslash}X} *{3}{@{\hspace{.9em}} >{\centering\arraybackslash}X} @{\hspace{.6em}}}
        \toprule
        
        \textbf{Dataset} & \textbf{Method} & \textbf{Compress.}$\uparrow$ & \textbf{PSNR}$\uparrow$ & \textbf{SSIM}$\uparrow$ & \textbf{LPIPS}$\downarrow$ \\
        
        \midrule

        \multirow{2}{*}{BAIR} & MAGVIT & 1024:1 & 24.1 & 0.834 & 0.156 \\
        & \textbf{\method} & 1024:1 & \textbf{28.8} & \textbf{0.950} & \textbf{0.013} \\

        \midrule

        \multirow{2}{*}{UCF} & MAGVIT & 1024:1 & 22.0 & 0.701 & 0.099 \\
        & \textbf{\method} & 1024:1 & \textbf{41.4} & \textbf{0.942} & \textbf{0.043} \\
        
        \midrule

        \multirow{3}{*}{K600} & Cosmos-DV & 256:1 & 36.9 & \textbf{0.973} & \textbf{0.031} \\
        & Cosmos-DV & \textbf{1024:1} & 31.4 & 0.928 & 0.048 \\
        & \textbf{\method} & \textbf{1024:1} & \textbf{42.9} & \underline{0.958} & \underline{0.034} \\

        \midrule

        \multirow{7}{*}{DAVIS} & CogVideoX & 256:1 & 31.7 & 0.860 & -- \\
        & VideoGPT & 64:1 & 32.2 & 0.850 & -- \\
        & Omni-VAE & 256:1 & 29.0 & 0.710 & -- \\
        & Omni-VQ & 256:1 & 28.4 & 0.712 & -- \\
        & Cosmos-DV & 256:1 & 33.0 & 0.818 & -- \\
        & Cosmos-CV & 256:1 & 35.3 & \textbf{0.890} & -- \\
        & \textbf{\method} & \textbf{1024:1} & \textbf{39.4} & \underline{0.888} & 0.112 \\

        \midrule

        \multirow{1}{*}{Avg.\ $\uparrow$\%} & \textbf{\method} & 1024:1 & \textbf{\hphantom{.}34\%} & \textbf{\hphantom{.}14\%} & \textbf{\hphantom{.}62\%} \\

        \bottomrule
    \end{tabularx}
    \caption{Quantitative results for video reconstruction. \method\ outperforms all baselines in cases when the compression ratio from the baseline is the same as ours, and is competitive in cases when the baseline compression ratio is lower. This demonstrates \method's ability to utilize reference frames to improve video reconstruction quality.}
    \label{tab:reconstruction_results}
\end{table}

\begin{table}[t]
    \centering
    \scriptsize
    \begin{tabularx}{\linewidth}{l | *{5}{>{\centering\arraybackslash}X} @{\hspace{.6em}}}
        \toprule
        
        \textbf{Method} & \textbf{Params}$\downarrow$ & \textbf{FVD}$\downarrow$ & \textbf{PSNR}$\uparrow$ & \textbf{SSIM}$\uparrow$ & \textbf{LPIPS}$\downarrow$ \\
        
        \midrule

        MAGVIT-B & 128M & 76 & -- & -- & -- \\
        MAGVIT-L & 464M & 62 & 19.3 & 0.787 & 0.123 \\
        
        \midrule
        
        \textbf{\method} & 128M & \textbf{57} & \textbf{23.6} & \textbf{0.878} & \textbf{0.037} \\
        
        \bottomrule
    \end{tabularx}
    \caption{Quantitative results for video generation. \method\ outperforms not only the base MAGVIT model (MAGVIT-B), but also its large-size variation (MAGVIT-L), in all video evaluation metrics. This demonstrates that \method\ can be used to improve the performance of downstream generation models.}
    \label{tab:generation_results}
\end{table}

\subsection{Application: Simple zero-shot video editing}
\begin{figure}
    \centering
    \includegraphics[width=\linewidth]{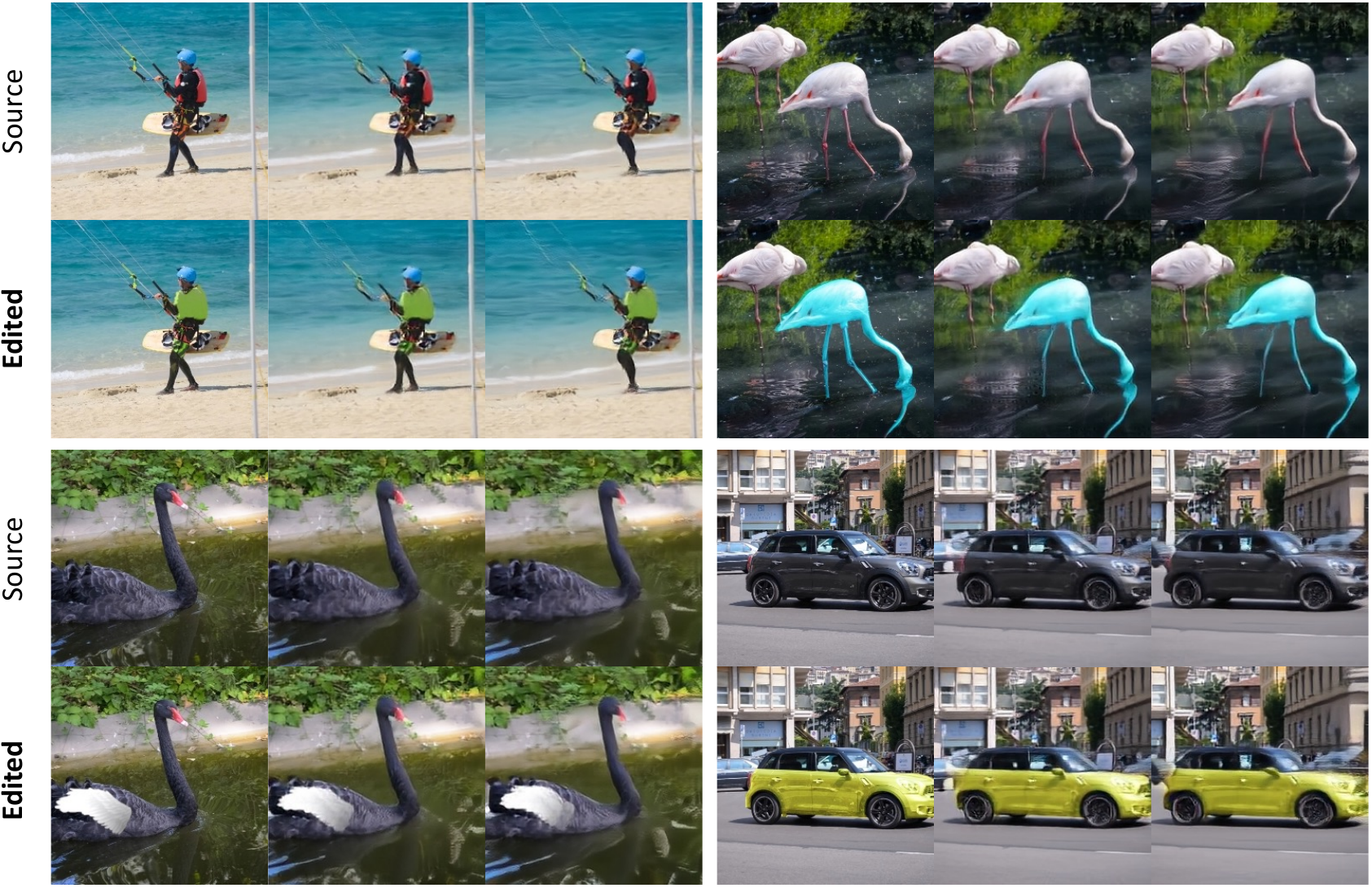}
    \caption{\method\ can perform simple video editing, such as changing the shape and color of a life vest, modifying the color of a flamingo and SUV car, and adding wings to a swan, without running expensive video diffusion models~\cite{fan2024videoshoplocalizedsemanticvideo} or atlas mapping~\cite{kasten2021layeredneuralatlasesconsistent}. This is achieved by using an edited reference frame for decoding.}
    \label{fig:editing}
\end{figure}

We showcase the zero-shot video editing capability of \method\ in \Cref{fig:editing}. By providing the model with an edited reference frame at decoding time, \method\ can perform simple video editing tasks such as changing the shape and color of a life vest, modifying the color of a flamingo and an SUV car, and adding wings to a swan. While simple, editing with \method\ only takes a single forward pass with less than a second, avoiding expensive video diffusion processes~\cite{fan2024videoshoplocalizedsemanticvideo,sun2024diffusionmodelbasedvideoediting,kara2023raverandomizednoiseshuffling,qi2023fatezerofusingattentionszeroshot} or atlas mapping~\cite{kasten2021layeredneuralatlasesconsistent,xu2022gmflowlearningopticalflow}.

\subsection{Ablation studies}
We ablate several design choices of \method. We experiment with changing the patch size to [1, 32, 32] (vs.\ original [4, 16, 16]), restricting attention to only the reference frame's tokens, and increasing the number of reference frames. These changes might plausibly be helpful as finer-grained temporal patches could improve representation, a more restricted attention mask might focus the model on relevant information, and additional reference frames could provide more context. These experiments are conducted on the K600 dataset, and the results are summarized in \Cref{tab:ablation_results}. The original design choices of \method\ are found to be optimal, as the larger patch size, more restricted attention mask, and increased reference frames do not meaningfully improve performance.

\begin{table}
    \centering
    \scriptsize
    \begin{tabularx}{.55\linewidth}{@{\hspace{.2em}} >{\centering\hsize=4.8em}X @{\hspace{.1em}} >{\centering\hsize=4.8em}X @{} | *{1}{>{\centering\arraybackslash}X} @{\hspace{4em}}}
        \toprule

        \multicolumn{2}{>{\hsize=9.6em}X}{\centering \textbf{$\Delta$}} & \textbf{L1 Error}$\downarrow$ ($\times 10^{-2}$) \\

        \midrule

        \multicolumn{2}{>{\hsize=9.6em}X |}{\centering ours} & 1.06 \\

        \midrule

        \multirow{1}{4.4em}{patch size} & [1, 32, 32] & 2.38 \\

        \midrule

        \multirow{1}{4.4em}{attn mask} & ref only & 1.63 \\

        \midrule

        \multirow{3}{4.4em}{ref frames} & 1 frame & 1.06 \\
        & 2 frames & 1.13 \\
        & 4 frames & 1.32 \\
        
        \bottomrule
    \end{tabularx}
    \label{tab:ablation_results}
    \caption{Ablations. We find that increasing patch size, employing more restricted attention mask, and increasing the number of reference frames do not meaningfully improve performance.}
    \vspace{-1.12em}
\end{table}

\section{Discussion and Limitations}
\label{sec:discussion}

We introduced \method, a reference-based tokenization approach for video generation. Our experiments show that \method\ produces higher quality reconstructions than state-of-the-art video tokenizers across various datasets, accurately reconstructing repeating concepts from the reference frame, while using higher compression ratios. We highlight \method's ability to reconstruct detailed text, faces, and patterns in moving scenes to maintain temporal consistency with the reference frame. Additionally, we demonstrate that \method\ can improve traditional video tokenizers in video generation tasks, enabling models to achieve state-of-the-art performance on frame prediction tasks without modifying the video generation architecture.

Despite its success, \method\ has limitations. First, it requires a reference frame, whereas an ideal model should integrate both reference-less and reference-based tokenization for longer videos. Reference frames are assumed to be present, but their utility and selection are not always clear. Future work could explore adaptive reference frame selection strategies, dynamically choosing the most relevant frame based on video content.

Second, \method\ may struggle with reconstructing content absent in the reference frame, such as new objects. The model should use the reference frame for existing objects while accurately reconstructing new objects. Recent advancements in tokenization methods have shown promise in reconstructing previously unseen objects, but further research is needed to combine reference-based tokenization with these advancements.

\newpage
{
    \small
    \bibliographystyle{ieeenat_fullname}
    \bibliography{main}
}

\end{document}